\documentclass[runningheads]{llncs}

\usepackage{amsmath}
\usepackage{amssymb}
\usepackage{graphicx}
\usepackage{booktabs}
\usepackage{multirow}
\usepackage{tabularx}
\usepackage{array}
\usepackage{placeins}
\usepackage{marvosym}
\usepackage{url}
\usepackage[hidelinks]{hyperref}

\hypersetup{
  pdftitle={TriageRA-CCF: Source-Side Clinical Confidence and Coverage Signals for Adaptive Rank Budgeting in Medical LLMs},
  pdfauthor={Shucan Ji, Yining Huang, Hongliang Guo}
}

\newcolumntype{Y}{>{\raggedright\arraybackslash}X}
\newcommand{\method}{TriageRA-CCF}

\title{\method: Source-Side Clinical Confidence and Coverage Signals for Adaptive Rank Budgeting in Medical LLMs}
\titlerunning{\method}
\author{Shucan Ji\inst{1} \and Yining Huang\inst{2} \and Hongliang Guo\inst{1}\textsuperscript{\Letter}}
\authorrunning{S. Ji et al.}
\institute{College of Computer Science, Sichuan University (SCU), Chengdu, China, 610064\\
\email{jishucan@stu.scu.edu.cn, guohongliang@scu.edu.cn}
\and
Meta Emergence Laboratory\\
\email{huangyining1987@gmail.com}}

\begin{document}
\maketitle

\begin{abstract}
Medical questions vary in confidence and source coverage, yet parameter-efficient adaptation normally applies one fixed LoRA rank to every input. We study input-conditioned rank budgeting, where a router selects 2, 4, or 8 active channels from one shared rank-16 basis. We propose \method, whose budget teacher combines source-model confidence, clinical metadata coverage, and a counterfactual close-miss proxy. To avoid target-benchmark model selection, the budget-cost coefficient is chosen on a deterministic held-out CMB source split and then frozen before four-benchmark evaluation. We further compare against fixed-rank LoRA at ranks 2, 4, 8, and 16 and repeat the strongest paired controls over three seeds. On Qwen3-8B, Full CCF obtains $69.059\pm0.496$ average accuracy versus $68.792\pm0.319$ for LoRA r8; all three paired differences are positive. On Llama3.1-8B, Full CCF obtains $58.758\pm0.505$ versus $58.614\pm0.401$ for LoRA r2, with one negative paired seed. Neither comparison is statistically significant with three seeds. The results support target-blind, input-dependent capacity control on Qwen while exposing backbone sensitivity, non-additive teacher signals, and the absence of demonstrated wall-clock savings.
\end{abstract}

\section{Introduction}

Medical multiple-choice question answering is a practical stress test for clinical knowledge in large language models (LLMs). CMB, CMExam, MedQA, and MedMCQA cover heterogeneous specialties, professions, and reasoning operations~\cite{wang2024cmb,liu2023cmexam,jin2020medqa,pal2022medmcqa}. Strong general models still require careful adaptation and validation before clinical use~\cite{singhal2022clinical,singhal2023expertlevel}. A common adaptation recipe is fixed-rank LoRA or a mixture of LoRA experts. This is simple, but it assumes that every question requires the same update capacity.

That assumption is poorly matched to medical exam data. Some questions are direct recall items whose correct option already has a large base-model margin; aggressive updates may add little or perturb a correct answer. Other questions are ambiguous, underrepresented in source training, or close to being solved by the base model. They may benefit from more adaptation capacity. The relevant decision is therefore not only what low-rank parameters to learn, but how much of that capacity to activate for each example.

\begin{figure}[!ht]
  \centering
  \includegraphics[width=\textwidth]{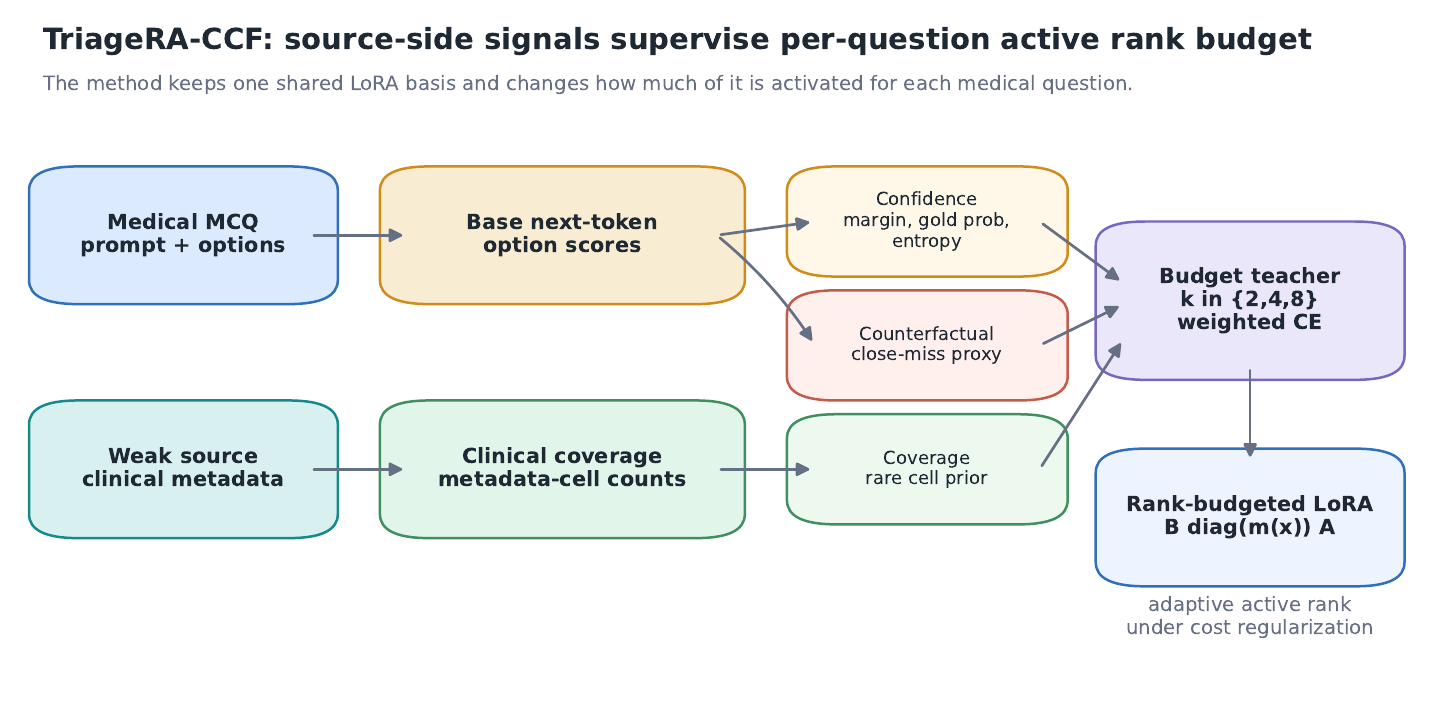}
  \caption{\method{} constructs source-side budget supervision from answer confidence, a counterfactual close-miss proxy, and clinical coverage. The learned router activates a small, medium, or large subset of channels inside one shared LoRA basis.}
  \label{fig:pipeline}
\end{figure}

We study adaptive budgeting inside one shared LoRA basis. The adapter gates rank channels in $B\operatorname{diag}(m(x))A$ and selects $k(x)\in\{2,4,8\}$ per input. A router trained only through answer loss can collapse to unstable choices, overuse high budgets, or learn a policy that does not transfer beyond the source data. We therefore introduce \method, where CCF denotes \textbf{confidence}, \textbf{clinical coverage}, and a \textbf{counterfactual proxy}. Confidence measures option margins and entropy. Coverage counts coarse source metadata cells. The close-miss proxy identifies wrong base answers whose gold option remains near the top. These source-only signals supervise the discrete budget router, as illustrated in Fig.~\ref{fig:pipeline}.

\suppressfloats[t]

Initial single-run evidence leaves three questions unresolved: whether an appropriate fixed rank is sufficient, whether the budget cost can be selected without looking at target benchmarks, and whether gains survive training variation. We address all three under a common protocol. Our contributions are:
\begin{itemize}
\item We formulate medical PEFT as input-conditioned active-rank budgeting within one shared adapter basis.
\item We introduce a source-side CCF teacher that uses confidence, metadata coverage, and close-miss structure without target labels.
\item We select the budget cost on a held-out source split, then freeze it for three-seed evaluation on four benchmarks.
\item We add fixed-rank LoRA with $r\in\{2,4,8,16\}$ and static-mask controls. We report paired variation and parameter counts, limiting the claim to modest backbone-dependent gains rather than universal superiority or measured speedup.
\end{itemize}

Table~\ref{tab:motivation} summarizes the design rationale. The rules are deliberately auditable rather than presented as uniquely optimal.

\suppressfloats[t]
\begin{table}[!ht]
\centering
\caption{Design motivations in \method.}
\label{tab:motivation}
\small
\begin{tabularx}{\textwidth}{lYY}
\toprule
Component & Design reason & Intended effect \\
\midrule
Shared rank basis & Avoids a bank of complete adapter experts. & Uses one budgeted adapter for all inputs. \\
Adaptive $k(x)$ & Medical questions differ in uncertainty and coverage. & Varies active capacity by input. \\
Confidence & Large option margins indicate easier source examples. & Discourages unnecessary high budgets. \\
Close miss & A wrong answer with gold near the top is plausibly repairable. & Prioritizes uncertain but actionable errors. \\
Clinical coverage & Rare metadata cells are underrepresented. & Raises the prior budget for under-covered question types. \\
Budget cost & A discrete router may always choose rank 8. & Regularizes the accuracy-capacity trade-off. \\
\bottomrule
\end{tabularx}
\end{table}

\section{Related Work}

\paragraph{Medical LLM evaluation.}
Medical QA benchmarks test both factual knowledge and clinical reasoning. PubMedQA targets biomedical research question answering~\cite{jin2019pubmedqa}; MedQA and MedMCQA collect exam-style multiple-choice questions in English~\cite{jin2020medqa,pal2022medmcqa}; and CMB and CMExam provide broad Chinese medical examinations across specialties and professions~\cite{wang2024cmb,liu2023cmexam}. Large medical LLM studies such as Med-PaLM and Med-PaLM 2 report strong progress but also emphasize calibration, uncertainty, and safety-sensitive evaluation~\cite{singhal2022clinical,singhal2023expertlevel}. Our work does not propose a new benchmark; instead, it asks whether a medical adapter should spend the same low-rank capacity on all benchmark items.

\paragraph{Parameter-efficient adaptation.}
LoRA freezes the base model and learns low-rank update matrices~\cite{hu2021lora}; QLoRA makes this practical for quantized LLM fine-tuning~\cite{dettmers2023qlora}; and DoRA decomposes weight magnitude and direction for stronger low-rank adaptation~\cite{liu2024dora}. Recent systems further adapt Qwen and Llama-family backbones~\cite{yang2025qwen3,grattafiori2024llama3}. We compare against LoRA and DoRA baselines under the same source data and training length. Unlike these fixed-rank methods, \method{} changes the active rank budget per question while retaining a single shared adapter basis.

\paragraph{Adaptive rank allocation.}
AdaLoRA allocates parameter budgets according to importance estimates of LoRA components~\cite{zhang2023adalora}. DyLoRA trains a range of ranks without retraining for each rank~\cite{valipour2023dylora}, and IncreLoRA incrementally allocates parameters across modules~\cite{zhang2023increlora}. These methods mainly address rank allocation across weights, modules, or deployment settings. Our focus is complementary: the rank budget is conditioned on the current medical question and supervised by source-side clinical uncertainty signals.

\paragraph{Mixture and routing over LoRA experts.}
Sparse mixture-of-experts models route examples or tokens to a subset of experts~\cite{shazeer2017moe,fedus2022switch}. LoRA-based expert mixtures, including MoELoRA, MoLE, MixLoRA, and MING-MOE, combine PEFT with expert routing~\cite{luo2024moelora,wu2024mole,li2024mixlora,liao2024mingmoe}. Medical MOE-LoRA variants use task-motivated expert gates for multi-task medical applications~\cite{liu2024moemeets}; LoRAHub dynamically composes a library of LoRA modules~\cite{huang2024lorahub}. \method{} differs in two ways. First, it does not train or select a bank of full LoRA experts; it varies active rank channels inside one adapter. Second, its supervision is not a task ID or expert label, but a source-side estimate of how much rank capacity the input should receive.

\paragraph{Calibration and uncertainty.}
Calibration work shows that neural confidence is often misaligned with correctness~\cite{guo2017calibration}, and test-time adaptation methods can use entropy or unlabeled statistics to adjust models~\cite{wang2021tent}. We use confidence only as one source-side teacher signal rather than as a deploy-time guarantee. Coverage and close-miss structure provide complementary hypotheses, but our ablations explicitly test and limit claims about their additivity.

\section{Methodology}
\label{sec:method}

\subsection{Problem Setting}

Let $x$ be a medical multiple-choice question with options and source answer $y$. Standard LoRA uses a fixed low-rank update
\begin{equation}
    W(x)h = W_0h + \frac{\alpha}{r}BAh,
\end{equation}
where $W_0$ is frozen and $A,B$ are trainable. We instead use an adaptive rank-budgeted update
\begin{equation}
    W(x)h = W_0h + \frac{\alpha}{r}B\operatorname{diag}(m(x))Ah,
\end{equation}
where $m(x)$ is a sparse nonnegative rank mask. The source-supervised budget branch determines the number of active entries in $m(x)$ for each input.

\subsection{Rank-Budgeted Adapter}

For each LoRA target module, rank logits are computed from the question representation,
\begin{equation}
    s(x)=f_{\mathrm{rank}}(h_x),
\end{equation}
where $h_x$ is a pooled hidden representation. The individual rank channels are treated as latent capacity units, while clinical metadata enters only through the coverage teacher below. Let $p_i(x)=\operatorname{softmax}(s(x))_i$. For a selected budget $k$, the forward pass keeps the top-$k$ probabilities:
\begin{equation}
    m_i^{(k)}(x)=
    \begin{cases}
    p_i(x)\tau_k(x), & i\in\operatorname{TopK}(p(x),k),\\
    0, & \text{otherwise},
    \end{cases}
\end{equation}
where $\tau_k(x)$ normalizes the selected mass so update magnitudes remain comparable across budgets. Rank-balance and entropy penalties discourage channel collapse and sharpen the distribution.

\subsection{Adaptive Budget Router}

The budget router chooses from $\mathcal K=\{2,4,8\}$. The small discrete set is easy to audit: rank 2 is conservative, rank 4 is the middle operating point, and rank 8 activates half of the stored rank-16 basis. From $h_x$, the router computes logits $b(x)=b_0+f_{\mathrm{budget}}(h_x)$ and probabilities $q(x)=\operatorname{softmax}(b(x)/T_b)$. Training uses a straight-through decision,
\begin{equation}
    \tilde q(x)=\operatorname{onehot}(\arg\max q(x))-\operatorname{sg}(q(x))+q(x),
\end{equation}
and combines candidate masks as $m(x)=\sum_j\tilde q_j(x)m^{(k_j)}(x)$. The expected-budget cost
\begin{equation}
    \mathcal L_{\mathrm{cost}}=\mathbb E_x\left[\frac{\sum_jq_j(x)k_j}{\max(\mathcal K)}\right]
\end{equation}
discourages always selecting rank 8.

\subsection{Source-Side CCF Budget Teacher}

Before adapter training, the frozen base model scores answer options by next-token probabilities over option letters. We also derive a coarse specialty or profession tag $r_x$ and a deterministic clinical-operation tag $o_x$ from source examples. These tags are used only to estimate source coverage. Three signals are computed:
\begin{itemize}
\item \textbf{Confidence}: top-1/top-2 margin, gold-option probability, and normalized option entropy.
\item \textbf{Counterfactual proxy}: a wrong prediction whose gold option remains in the top two and close to the predicted option.
\item \textbf{Clinical coverage}: the frequency of source metadata cells formed by the coarse route and operation tags.
\end{itemize}
They produce a teacher class $z(x)\in\{0,1,2\}$ for active ranks 2, 4, and 8 and a weight $w(x)$. High-budget labels are assigned for close misses or strong uncertainty, medium labels for moderate uncertainty or rare cells, and low labels otherwise. The source-only restriction prevents the teacher from becoming a hidden target-domain selector. Table~\ref{tab:teacher-rule} records the deterministic rule. Thresholds discretize uncertainty and are not claimed to be theoretically unique.

\suppressfloats[t]
\begin{table}[!ht]
\centering
\caption{Source-side CCF teacher rule. Entropy is normalized by the option count; the rare-cell threshold is the 45th percentile of non-empty source-cell counts.}
\label{tab:teacher-rule}
\scriptsize
\setlength{\tabcolsep}{4pt}
\begin{tabularx}{\textwidth}{lY}
\toprule
Item & Setting \\
\midrule
Budget classes & Low, medium, and high active ranks: $\{2,4,8\}$ \\
Confidence features & Top-1/top-2 margin, gold probability, normalized entropy \\
Thresholds & Low margin 0.08, medium margin 0.22, low gold probability 0.22, medium entropy 0.68, high entropy 0.84 \\
Close miss & Base prediction wrong, gold in top two, and margin $\leq0.22$ or gold probability $\geq0.22$ \\
High budget & Close miss, wrong with strong uncertainty, or entropy $\geq0.84$ \\
Medium budget & Margin $\leq0.22$, entropy $\geq0.68$, or rare source cell \\
Teacher weights & 1.15 for high, 1.0 by default, 0.9 for rare-cell-only medium \\
\bottomrule
\end{tabularx}
\end{table}

\subsection{Training and Source-Only Selection}

The training objective is
\begin{equation}
\mathcal L=\mathcal L_{\mathrm{ans}}+\beta\mathcal L_{\mathrm{balance}}+\eta\mathcal L_{\mathrm{entropy}}+\rho\mathcal L_{\mathrm{cost}}+\mu\mathcal L_{\mathrm{teacher}},
\end{equation}
where
\begin{equation}
\mathcal L_{\mathrm{teacher}}=-\frac{\sum_xw(x)\log q_{z(x)}(x)}{\max(1,\sum_xw(x))}.
\end{equation}
We fix $\beta=0.001$, $\eta=0.01$, $\mu=0.2$, and $T_b=0.7$. The only selected coefficient is $\rho\in\{0,0.02,0.05\}$. It is chosen separately for each backbone using source-validation accuracy, with lower active rank and then lower $\rho$ as tie breakers. Target benchmark scores are not consulted. After selection, the chosen $\rho$ is frozen and the model is retrained on all source examples for final evaluation.

\section{Experimental Setup}
\label{sec:experiments}

\subsection{Data and Evaluation Protocol}

All methods use CMB source data, maximum sequence length 768, 400 update steps, micro-batch size 1, gradient accumulation 8, and answer-only supervision. For selection, the 4,200 source examples are deterministically divided into 3,360 training and 840 validation examples with seed 20260820, stratified by coarse route, weak operation, and answer label; source UIDs do not overlap. Final models are retrained on all 4,200 examples. Evaluation is frozen across CMB eval4149 (4,149 Chinese questions), CMExam full (6,811 Chinese questions), MedQA test700, and MedMCQA val700. Accuracy is computed from generated answer letters. The four-dataset average is an unweighted macro-average.

\suppressfloats[t]
\begin{table}[!ht]
\centering
\caption{Unified evaluation protocol retained across all comparisons.}
\label{tab:protocol}
\small
\begin{tabularx}{\textwidth}{lrrY}
\toprule
Dataset & Size & Language & Role \\
\midrule
CMB eval4149 & 4149 & Chinese & in-domain medical exam evaluation \\
CMExam full & 6811 & Chinese & shifted Chinese medical exam evaluation \\
MedQA test700 & 700 & English & shifted English medical exam evaluation \\
MedMCQA val700 & 700 & English & shifted English multi-subject medical QA evaluation \\
\bottomrule
\end{tabularx}
\end{table}

\FloatBarrier

\subsection{Backbones and Baselines}

We evaluate Qwen3-8B~\cite{yang2025qwen3} and Llama3.1-8B-Instruct~\cite{grattafiori2024llama3}. \method{} stores rank 16 with alpha 32 and targets \texttt{q/k/v/o/gate/up/down}. Primary controls are ordinary LoRA ranks 2, 4, 8, and 16 with alpha $2r$, selected independently per backbone from the fixed-rank sweep. We also evaluate static masks with $k\in\{2,8\}$ inside the same stored rank-16 routing basis. Additional external controls are LoRA r24/a48, DoRA r24/a48, and MoELoRA with four rank-8 experts and top-2 routing. Every paired run uses seeds 13, 42, and 2026 and exactly the same data, update count, effective batch size, and evaluator.

\suppressfloats[t]
\begin{table}[!ht]
\centering
\caption{Main implementation settings. All methods use the same source examples, sequence length, update count, and effective batch size.}
\label{tab:impl}
\small
\begin{tabularx}{\textwidth}{lY}
\toprule
Item & Setting \\
\midrule
Training source & CMB train, maximum 4,200 examples \\
Optimization length & 400 updates, micro-batch 1, gradient accumulation 8 \\
Sequence length & Maximum 768 tokens, left truncation when needed \\
\method{} adapter & Rank 16, alpha 32, dropout 0.05, targets \texttt{q/k/v/o/gate/up/down} \\
Adaptive budgets & Active rank choices $\{2,4,8\}$, budget temperature 0.7 \\
Regularization & Rank balance 0.001, entropy 0.01, teacher coefficient 0.2 \\
Budget cost candidates & $\rho\in\{0,0.02,0.05\}$; selected on CMB source validation \\
External PEFT baselines & LoRA r24/a48, DoRA r24/a48, MoELoRA 4 experts/top-2/r8 \\
\bottomrule
\end{tabularx}
\end{table}

\subsection{Capacity Accounting}

Active rank is a routing diagnostic, not a measured latency result. The Qwen and Llama CCF models respectively train 69.31M and 66.00M parameters (1.448\% and 1.433\% of the adapted models), including the stored rank-16 basis and routers. Their strongest fixed controls train 21.82M parameters for Qwen LoRA r8 and 5.24M for Llama LoRA r2. Thus, lower active rank does not imply lower stored or trainable parameter count. Moreover, the current masked implementation does not use a fused sparse-rank kernel, so we make no claim of wall-clock, throughput, memory, or FLOP reduction.

\suppressfloats[t]
\begin{table}[!ht]
\centering
\caption{Capacity accounting for the primary paired comparison. Trainable parameters include the budget and rank routers for Full CCF.}
\label{tab:capacity}
\small
\begin{tabular}{llrrr}
\toprule
Backbone & Method & Stored rank & Active rank & Trainable (M) \\
\midrule
Qwen3-8B & LoRA r8 & 8 & 8 & 21.823 \\
Qwen3-8B & Full CCF & 16 & 4.036 & 69.311 \\
Llama3.1-8B & LoRA r2 & 2 & 2 & 5.243 \\
Llama3.1-8B & Full CCF & 16 & 4.928 & 66.001 \\
\bottomrule
\end{tabular}
\end{table}

\section{Results and Analysis}
\label{sec:results}

\subsection{Source-Selected Multi-Seed Results}

Table~\ref{tab:source-selection} reports the only model-selection stage. The 840-example CMB source-validation split selects $\rho=0.05$ for Qwen and $\rho=0.02$ for Llama before any target benchmark is evaluated. The selected setting is then retrained on all 4,200 source examples. This ordering differs from the target-benchmark sensitivity screen in Sect.~\ref{sec:cost-screen}, demonstrating why the independent split matters.

\suppressfloats[t]
\begin{table}[!ht]
\centering
\caption{Budget-cost selection using CMB source validation only. Bold rows are frozen before target evaluation.}
\label{tab:source-selection}
\small
\begin{tabular}{llrr@{\qquad}llrr}
\toprule
Backbone & $\rho$ & Source val. & Active & Backbone & $\rho$ & Source val. & Active \\
\midrule
Qwen3-8B & .00 & 80.595 & 3.942 & Llama3.1-8B & .00 & 65.119 & 4.924 \\
Qwen3-8B & .02 & 80.833 & 3.978 & Llama3.1-8B & \textbf{.02} & \textbf{66.071} & \textbf{4.895} \\
Qwen3-8B & \textbf{.05} & \textbf{81.071} & \textbf{3.976} & Llama3.1-8B & .05 & 64.881 & 4.951 \\
\bottomrule
\end{tabular}
\end{table}

Table~\ref{tab:multiseed} is the primary target result. On Qwen, Full CCF improves the four-benchmark average by 0.267 points over LoRA r8, with a positive mean difference on every benchmark. On Llama, the average improvement over LoRA r2 is 0.144 points, but CMExam and MedQA favor the fixed control.

\suppressfloats[t]
\begin{table}[!ht]
\centering
\caption{Three-seed target evaluation after source-only selection. Dataset columns are means; Avg. reports mean $\pm$ sample standard deviation.}
\label{tab:multiseed}
\scriptsize
\setlength{\tabcolsep}{2.7pt}
\begin{tabular}{llrrrrrr}
\toprule
Backbone & Method & CMB & CMExam & MedQA & MedMCQA & Avg. & Active \\
\midrule
\multirow{2}{*}{Qwen3-8B}
 & LoRA r8 & 78.099 & 76.308 & 63.048 & 57.714 & $68.792\pm0.319$ & 8 \\
 & Full CCF & \textbf{78.123} & \textbf{76.494} & \textbf{63.571} & \textbf{58.048} & $\mathbf{69.059\pm0.496}$ & 4.036 \\
\midrule
\multirow{2}{*}{Llama3.1-8B}
 & LoRA r2 & 56.118 & \textbf{56.051} & \textbf{62.667} & 59.619 & $58.614\pm0.401$ & 2 \\
 & Full CCF & \textbf{56.222} & 55.763 & 62.429 & \textbf{60.619} & $\mathbf{58.758\pm0.505}$ & 4.928 \\
\bottomrule
\end{tabular}
\end{table}

The Qwen paired differences for seeds 13, 42, and 2026 are $+0.489$, $+0.020$, and $+0.291$ points; Llama gives $+0.623$, $-0.784$, and $+0.594$. Two-sided paired $t$-tests yield $p=0.189$ and $p=0.785$, with 95\% confidence intervals $[-0.318,0.852]$ and $[-1.853,2.142]$. Neither comparison is statistically significant with three pairs. Complete per-seed scores appear in Appendix~\ref{app:seed-results}.

\FloatBarrier

\subsection{Single-Run External PEFT Context}

\begin{figure}[!ht]
  \centering
  \includegraphics[width=.82\textwidth]{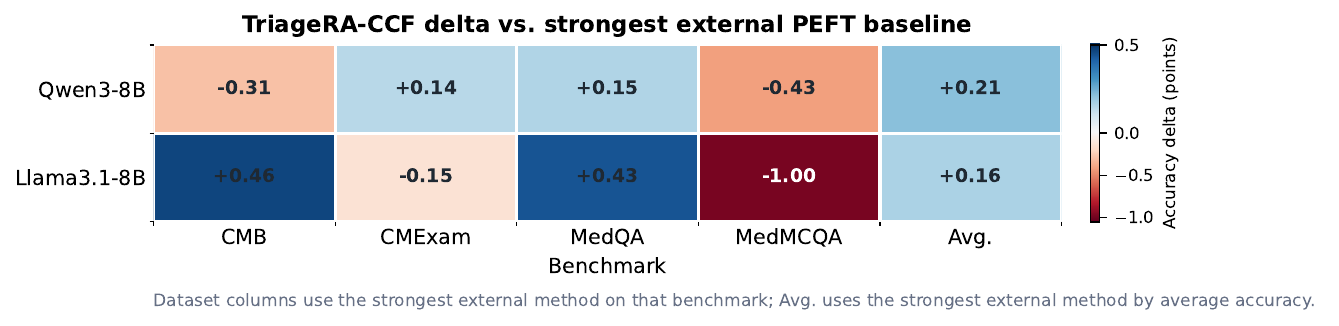}
  \caption{Per-benchmark accuracy delta of the screened CCF setting relative to the strongest external PEFT baseline. Dataset columns use the strongest external method on that benchmark; Avg. uses the strongest external method by average accuracy.}
  \label{fig:external-screen}
\end{figure}

Table~\ref{tab:external-screen} and Fig.~\ref{fig:external-screen} retain the original single-run external comparison. The CCF rows use the strongest operating point from the target-benchmark cost screen, so they are contextual sensitivity results rather than the primary target-blind evidence. Under this screen, CCF averages 68.79 on Qwen and 58.51 on Llama, 0.21 and 0.16 points above the strongest external average. In contrast, the strict source-selected seed-42 models score 68.49 and 58.22, below the best external average by 0.09 and 0.13 points.

\suppressfloats[t]
\begin{table}[!ht]
\centering
\caption{Original single-run external PEFT comparison under the matched CMB-source protocol. ``Gain'' uses the strongest external average for each backbone.}
\label{tab:external-screen}
\scriptsize
\setlength{\tabcolsep}{4pt}
\begin{tabular}{llrrrrrr}
\toprule
Backbone & Method & CMB & CMExam & MedQA & MedMCQA & Avg. & Gain \\
\midrule
\multirow{4}{*}{Qwen3-8B}
& LoRA r24/a48 & 78.09 & 75.82 & 62.57 & 57.86 & 68.58 & -- \\
& DoRA r24/a48 & 78.69 & 75.94 & 62.00 & 56.71 & 68.34 & -- \\
& MoELoRA 4e/top2/r8 & 77.95 & 75.32 & 63.14 & 57.29 & 68.42 & -- \\
& Full CCF, $\rho=.02$ & 78.38 & 76.08 & 63.29 & 57.43 & \textbf{68.79} & \textbf{+.21} \\
\midrule
\multirow{4}{*}{Llama3.1-8B}
& LoRA r24/a48 & 55.89 & 54.49 & 61.14 & 59.57 & 57.77 & -- \\
& DoRA r24/a48 & 55.75 & 54.97 & 61.86 & 60.57 & 58.29 & -- \\
& MoELoRA 4e/top2/r8 & 55.87 & 55.97 & 61.57 & 60.00 & 58.35 & -- \\
& Full CCF, $\rho=.05$ & 56.35 & 55.82 & 62.29 & 59.57 & \textbf{58.51} & \textbf{+.16} \\
\bottomrule
\end{tabular}
\end{table}

\suppressfloats[t]

\FloatBarrier

\subsection{Target-Benchmark Budget-Cost Sensitivity}
\label{sec:cost-screen}

The original cost screen in Table~\ref{tab:cost} is preserved to show sensitivity. Qwen is flat around $\rho\in\{0,.02\}$, while Llama is strongest at $.05$. Average active rank remains near four for Qwen and five for Llama rather than collapsing to the minimum. Because target accuracy appears in this screen, it is not used to choose the primary multi-seed configuration.

\suppressfloats[t]
\begin{table}[!ht]
\centering
\caption{Original target-benchmark budget-cost sensitivity. Active rank is averaged over the four evaluation benchmarks.}
\label{tab:cost}
\scriptsize
\setlength{\tabcolsep}{3.3pt}
\begin{tabular}{llrrrrrr}
\toprule
Backbone & $\rho$ & CMB & CMExam & MedQA & MedMCQA & Avg. & Active \\
\midrule
\multirow{3}{*}{Qwen3-8B}
& .00 & 78.31 & 75.83 & 63.71 & 57.29 & \textbf{68.79} & 4.106 \\
& .02 & 78.38 & 76.08 & 63.29 & 57.43 & \textbf{68.79} & 4.118 \\
& .05 & 78.50 & 75.95 & 63.43 & 57.00 & 68.72 & 4.136 \\
\midrule
\multirow{3}{*}{Llama3.1-8B}
& .00 & 56.06 & 55.69 & 61.29 & 59.86 & 58.22 & 4.921 \\
& .02 & 56.16 & 55.26 & 62.00 & 59.43 & 58.21 & 5.004 \\
& .05 & 56.35 & 55.82 & 62.29 & 59.57 & \textbf{58.51} & 4.986 \\
\bottomrule
\end{tabular}
\end{table}

\suppressfloats[t]
\begin{figure}[!ht]
  \centering
  \includegraphics[width=.86\textwidth]{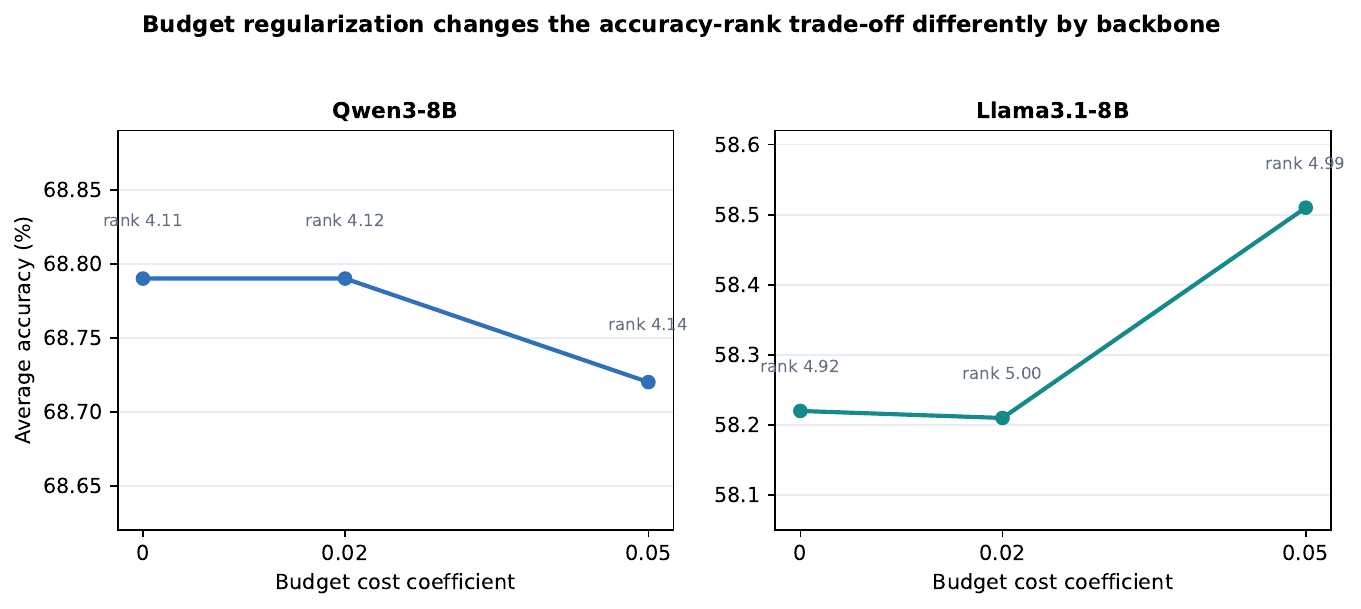}
  \caption{Original budget-cost sensitivity. Qwen is stable around $\rho\in\{0,.02\}$, while Llama benefits from the stronger $.05$ cost in the target screen.}
  \label{fig:cost}
\end{figure}

\FloatBarrier

\subsection{Fixed-Rank and Static-Budget Controls}

Table~\ref{tab:fixed-controls} asks whether a well-chosen static rank is sufficient. The trend is non-monotonic: Qwen peaks at ordinary LoRA r8, while Llama peaks at r2. Static masks are strong on Qwen. In the shared seed-42 diagnostic, source-selected Full CCF does not beat every static alternative; the positive Qwen evidence instead comes from its directionally consistent three-seed comparison.

\suppressfloats[t]
\begin{table}[!ht]
\centering
\caption{Single-seed fixed-rank and static-mask diagnostic. Entries are four-benchmark average accuracy; all rows use seed 42.}
\label{tab:fixed-controls}
\small
\begin{tabular}{lrrl}
\toprule
Control & Qwen3-8B & Llama3.1-8B & Capacity setting \\
\midrule
Ordinary LoRA r2 & 68.01 & \textbf{59.02} & fixed rank 2 \\
Ordinary LoRA r4 & 68.04 & 58.81 & fixed rank 4 \\
Ordinary LoRA r8 & \textbf{68.41} & 58.57 & fixed rank 8 \\
Ordinary LoRA r16 & 68.36 & 58.37 & fixed rank 16 \\
Static mask k2 & 68.74 & 57.79 & stored r16, active 2 \\
Static mask k8 & \textbf{68.74} & \textbf{58.60} & stored r16, active 8 \\
Source-selected CCF & 68.49 & 58.22 & stored r16, adaptive $\{2,4,8\}$ \\
\bottomrule
\end{tabular}
\end{table}

Reproduction is close across queues: Qwen r8 changes from 68.41 here to 68.47 for seed 42 in the paired queue, and Llama r2 changes from 59.02 to 59.01. This supports consistency of the baseline implementation while ruling out a universal adaptive-rank advantage.

\subsection{Teacher Signal Component Ablation}

\begin{figure}[!ht]
  \centering
  \includegraphics[width=.86\textwidth]{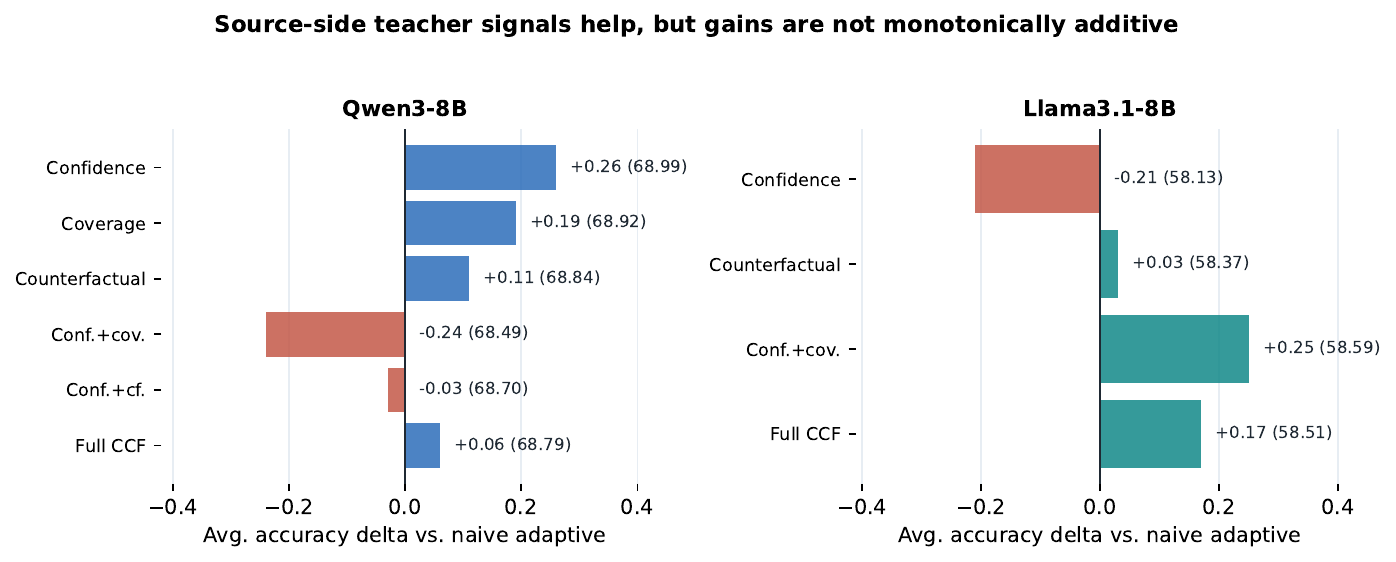}
\caption{Average accuracy change of teacher variants relative to the naive adaptive router. Components are useful, but the combined teacher is not strictly monotonic across backbones.}
  \label{fig:components}
\end{figure}

The original component screen is retained in Table~\ref{tab:components} and Fig.~\ref{fig:components}. On Qwen, confidence-only, coverage-only, counterfactual-only, and full CCF improve over the naive router, with confidence-only highest. On Llama, confidence+coverage is strongest and full CCF remains close. The Llama screen uses $\rho=.02$ for the naive router and $.05$ for teacher variants, so it identifies useful signals and possible interference rather than a controlled causal decomposition.

\suppressfloats[t]
\begin{table}[!ht]
\centering
\caption{Original teacher signal component screen. Signals are useful, but full CCF is not monotonically best on every backbone.}
\label{tab:components}
\scriptsize
\setlength{\tabcolsep}{1.8pt}
\begin{tabular}{llrrrrrrr}
\toprule
Backbone & Teacher variant & $\rho$ & CMB & CMExam & MedQA & MedMCQA & Avg. & Active \\
\midrule
Qwen3-8B & Naive adaptive & .02 & 78.72 & 75.92 & 62.71 & 57.57 & 68.73 & 4.013 \\
Qwen3-8B & Confidence only & .02 & 77.97 & 76.01 & 64.57 & 57.43 & \textbf{68.99} & 4.119 \\
Qwen3-8B & Coverage only & .02 & 78.40 & 75.86 & 63.71 & 57.71 & 68.92 & 4.008 \\
Qwen3-8B & Counterfactual only & .02 & 78.50 & 76.13 & 63.00 & 57.71 & 68.84 & 4.106 \\
Qwen3-8B & Confidence+coverage & .02 & 77.92 & 75.60 & 62.86 & 57.57 & 68.49 & 4.068 \\
Qwen3-8B & Confidence+counterfactual & .02 & 78.57 & 75.94 & 63.29 & 57.00 & 68.70 & 4.134 \\
Qwen3-8B & Full CCF & .02 & 78.38 & 76.08 & 63.29 & 57.43 & 68.79 & 4.118 \\
\midrule
Llama3.1-8B & Naive adaptive & .02 & 56.25 & 55.40 & 61.29 & 60.43 & 58.34 & 5.039 \\
Llama3.1-8B & Confidence only & .05 & 56.47 & 55.78 & 61.43 & 58.86 & 58.13 & 4.980 \\
Llama3.1-8B & Counterfactual only & .05 & 56.23 & 56.12 & 61.14 & 60.00 & 58.37 & 5.052 \\
Llama3.1-8B & Confidence+coverage & .05 & 56.30 & 55.78 & 62.43 & 59.86 & \textbf{58.59} & 4.976 \\
Llama3.1-8B & Full CCF & .05 & 56.35 & 55.82 & 62.29 & 59.57 & 58.51 & 4.986 \\
\bottomrule
\end{tabular}
\end{table}

\suppressfloats[t]

\FloatBarrier

\subsection{Per-Benchmark Observations}

Benchmark effects are non-uniform. In the original screen, Qwen benefits most on CMExam and MedQA relative to strong external baselines, while Llama benefits most on CMB and MedQA. MedMCQA remains difficult: CCF is competitive but does not dominate every baseline. The stricter multi-seed comparison is consistent on Qwen but mixed on Llama, reinforcing that source-side clinical uncertainty is an inductive bias rather than a guarantee of improvement.

\FloatBarrier

\section{Limitations and Ethical Considerations}

\paragraph{Small and backbone-dependent effects.}
The average paired gains are 0.267 and 0.144 points and are not significant with three seeds. Qwen is directionally consistent, whereas Llama is not. We therefore do not claim universal superiority over fixed-rank LoRA.

\paragraph{Teacher heuristics and validation scope.}
The CCF thresholds are auditable source-side heuristics rather than theoretically optimal boundaries. Full CCF is not always better than a subset. Budget cost is now selected target-blind, but only one source split and three candidates are studied.

\paragraph{Parameters and runtime.}
The method stores rank 16 plus routers and trains more parameters than the strongest r8/r2 controls. Active rank describes the learned policy, not actual sparse execution. No fused kernel, latency, throughput, memory, or FLOP advantage has been demonstrated.

\paragraph{Data and metadata.}
All models train on 4,200 CMB examples. Specialty, profession, and operation tags are weak metadata rather than expert annotations. Results may change with larger corpora, clinical notes, or multilingual source mixtures.

\paragraph{Clinical use.}
Evaluation is limited to multiple-choice examinations. The system must not be used for diagnosis, treatment, or patient-facing advice without separate clinical validation, calibrated uncertainty, and human oversight.

\section{Conclusion}

We presented \method, a source-supervised router that selects input-dependent active rank inside one shared LoRA basis. A held-out source split removes target-benchmark selection of the budget cost, while fixed-rank sweeps and three-seed paired controls test whether adaptation gains exceed choosing a suitable static rank. Qwen shows a small, directionally consistent improvement over LoRA r8; Llama shows a smaller and seed-dependent mean improvement over LoRA r2. The evidence supports adaptive capacity control as a useful inductive bias on Qwen, not a universal accuracy or efficiency advantage. More seeds, learned source-side teachers, and sparse kernels are needed to establish statistical and practical benefits.

\paragraph{Disclosure of Interests.}
The authors have no competing interests to declare that are relevant to the content of this article.

\appendix

\section{Per-Seed Target Results}
\label{app:seed-results}

Table~\ref{tab:per-seed} gives every paired run used in Table~\ref{tab:multiseed}. All values are accuracies in percentage points. The adapter cost is fixed at $\rho=0.05$ for Qwen and $\rho=0.02$ for Llama before these runs are evaluated.

\suppressfloats[t]
\begin{table}[!ht]
\centering
\caption{Complete target-benchmark results for the paired multi-seed comparison.}
\label{tab:per-seed}
\scriptsize
\setlength{\tabcolsep}{2.7pt}
\begin{tabular}{lllrrrrrr}
\toprule
Backbone & Seed & Method & CMB & CMExam & MedQA & MedMCQA & Avg. & Active \\
\midrule
\multirow{6}{*}{Qwen3-8B}
 & 13 & LoRA r8 & 77.778 & 76.567 & 62.714 & 58.143 & 68.801 & 8 \\
 & 13 & Full CCF & 78.332 & 76.685 & 63.286 & 58.857 & 69.290 & 3.973 \\
 & 42 & LoRA r8 & 78.404 & 75.760 & 63.143 & 56.571 & 68.470 & 8 \\
 & 42 & Full CCF & 78.236 & 76.009 & 62.429 & 57.286 & 68.490 & 4.060 \\
 & 2026 & LoRA r8 & 78.115 & 76.597 & 63.286 & 58.429 & 69.107 & 8 \\
 & 2026 & Full CCF & 77.802 & 76.788 & 65.000 & 58.000 & 69.397 & 4.075 \\
\midrule
\multirow{6}{*}{Llama3.1-8B}
 & 13 & LoRA r2 & 55.941 & 56.306 & 61.286 & 59.286 & 58.205 & 2 \\
 & 13 & Full CCF & 56.375 & 55.792 & 62.000 & 61.143 & 58.827 & 4.985 \\
 & 42 & LoRA r2 & 56.712 & 56.027 & 63.571 & 59.714 & 59.006 & 2 \\
 & 42 & Full CCF & 56.255 & 55.777 & 61.571 & 59.286 & 58.222 & 4.941 \\
 & 2026 & LoRA r2 & 55.700 & 55.821 & 63.143 & 59.857 & 58.630 & 2 \\
 & 2026 & Full CCF & 56.038 & 55.719 & 63.714 & 61.429 & 59.225 & 4.857 \\
\bottomrule
\end{tabular}
\end{table}

These rows show why averages alone are insufficient. Qwen CCF gains on MedQA for seeds 13 and 2026 but loses 0.714 points for seed 42, while improvements on other datasets are smaller. Llama's seed-42 decline is distributed across all four benchmarks. The method therefore changes which items are solved rather than uniformly shifting every dataset.

\FloatBarrier
\bibliographystyle{splncs04}
\bibliography{references}

\end{document}